\documentclass[12pt]{article}
\usepackage[margin=1in]{geometry}
\usepackage{setspace}
\usepackage{amsmath,amssymb,amsthm}
\usepackage{graphicx}
\usepackage{booktabs}
\usepackage{multirow}
\usepackage{array}
\usepackage{xcolor}
\usepackage[round,authoryear]{natbib}
\let\cite\citep
\usepackage{url}
\usepackage{textcomp}
\usepackage[hidelinks]{hyperref}

\newtheorem{finding}{Finding}
\begin{document}
\onehalfspacing
\title{Context-Augmented, Session-Calibrated Blood Pressure Tracking from Photoplethysmography}
\author{Farouk Ganiyu Adewumi and Timothy Oladunni\\
\normalsize Department of Computer Science, Morgan State University, Baltimore, MD 21251, USA\\
\normalsize \texttt{fagan1@morgan.edu}; \texttt{timothy.oladunni@morgan.edu}}
\date{}
\maketitle
\begin{abstract}
\textbf{Objective.} To distinguish the contributions of acquisition context and photoplethysmography (PPG) to session-calibrated blood pressure (BP) tracking.

\textbf{Approach.} The original comparison used three repeats of five-fold participant-level cross-validation in 653 AuroraBP participants, excluding calibration-labelled prediction targets. Matched analyses separated context and PPG and removed elapsed-time terms. Later exploratory follow-ups retained PPG, accelerometer signals and three initial cuff readings. Separate studies evaluated pulse-rate fusion and an intraoperative VitalDB model.

\textbf{Main results.} The original wearable-state model achieved participant-level mean absolute errors (MAEs) of $7.697$/$6.139$~mmHg for systolic/diastolic BP (SBP/DBP). Added context reduced MAE by $0.436$/$0.126$~mmHg (97.5\% intervals $0.305$--$0.573$/$0.053$--$0.200$). In the matched factorial without the external PPG prior, PPG improved on calibration, time and protocol-provided posture/activity by $0.214$/$0.135$~mmHg; removing time left the DBP interval including zero. Standard PPG descriptors contributed, but additional nonlinear/attractor benefit was not established. In a separate follow-up series, a sensor-history ensemble achieved $7.299$/$5.445$~mmHg versus its static-rich reference's $7.363$/$5.470$, small gains of $0.064$/$0.025$~mmHg. Larger DBP errors did not improve uniformly. Quality-weighted fusion reduced pulse-rate MAE from $3.985$ to $3.690$ beats per minute. VitalDB MAEs remained $15.375$/$8.833$~mmHg despite small PPG gains.

\textbf{Significance.} The contribution is a calibrated evaluation framework with explicit controls, reproducible feature-cache-to-results analyses and retained negative findings. The sensor-history gain is exploratory after repeated dataset reuse. Independently estimated wearable states, complete raw-signal reproduction of the supplied richer features, external transfer and clinical adequacy remain unestablished.
\end{abstract}
\section{Introduction}
\label{sec:intro}
\subsection{Motivation}
Accurate blood pressure (BP) measurement is essential for identifying and managing hypertension. More frequent monitoring could characterize changes outside intermittent measurements, but repeated cuff inflation creates practical burdens. Optical sensing motivates research into less intrusive tracking; its potential clinical value depends on accuracy during changing activity and posture and on stability after calibration. Current guidance identifies important gaps in cuffless-device validation and clinical interpretation \cite{cohen2026cuffless,recommendations2024}. A clinically relevant first question is whether PPG contributes information about subsequent BP changes beyond the initial cuff anchor and acquisition context. This retrospective study examines that question within controlled recording protocols; it does not evaluate treatment decisions or clinical outcomes.

Participant overlap, calibration burden and protocol shortcuts can obscure this question. Record-level splits can inflate accuracy when correlated measurements from one participant enter both training and testing \cite{bpbenchmark2023}. Three initial cuff readings also represent a different calibration burden from one reading or repeated access to later reference values. Pigmentation-related optical differences motivate intervention testing, but small subgroups and nonsignificant contrasts cannot establish equitable performance \cite{fitzpatrick2024,jmir2024,phantom2026}.
We audit anchoring, context, and pigmentation-related interventions using recorded Fitzpatrick labels. Candidate interventions and context arms were frozen before repeated participant-level validation, after earlier development on AuroraBP. Because a fixed holdout was inspected across successive ablations, it is retained as historical development evidence. The primary repeated analysis confines feature ranking and fitting to outer training participants but remains internal validation. Secondary negative results are reported without interpreting them as proof of equity, equivalence, or general failure of pigmentation-aware modeling.
\subsection{Study Questions and Contributions}
The primary retrospective question is whether wearable-state context improves calibrated SBP/DBP tracking beyond the shared model core. Matched extensions isolate PPG/context and remove time. Secondary analyses examine calibration dose, nonlinear augmentation and pigmentation interventions; historical results quantify dependence on recent-cuff anchoring. Independent pulse-fusion and separately fitted VitalDB studies answer distinct questions. A later PPG/accelerometer series tests median prediction, direct BP targets and sensor history under the same three-reading burden. Negative findings are retained. Hashes, environments, folds and numerical checks support computational reproduction, while repeated dataset reuse precludes prospective confirmation.
\section{Related Work}
\label{sec:related}
\subsection{PPG-Based Cuffless Blood Pressure Estimation}
Cuffless BP estimation from PPG includes pulse-transit-time methods, waveform features, and end-to-end networks \cite{elgendi2019ppg,mukkamala2015toward}. Calibration-free methods avoid subject-specific references, whereas calibrated approaches use one or more cuff readings to model subsequent change \cite{calfree2023,onetimecal2023,avct2026}. Evaluation guidance emphasizes participant separation, calibration burden, and careful use of AAMI/ISO~81060-2 and BHS criteria \cite{recommendations2024}.
\subsection{Nonlinear and Information-Theoretic Feature Representations}
The feature bank combines waveform moments with delay-coordinate reconstruction, recurrence analysis, sample entropy, permutation entropy, and spectral measures \cite{takens1981,eckmann1987,marwan2007,richman2000,bandt2002}. These established descriptors have been applied to physiological and arterial pulse signals \cite{allen2007ppg,attractorpulse2022}. Recent author-developed formulations motivated the feature grouping and terminology \cite{cst2026,avct2026,adt2026}, but are not treated as validated physiological laws. The matched ablation therefore tests the empirical contribution of the combined nonlinear/attractor family rather than using prediction as evidence for the underlying theories.
\subsection{Skin Pigmentation and PPG-Derived Vitals}
Pigmentation can affect optical measurements, motivating separate evaluation of signal quality and downstream error \cite{jmir2024,fitzpatrick2024,phantom2026,montecarlo2026}. Existing field, polarization, and gain-correction studies support testing optical interventions but do not establish improved BP tracking \cite{equity2026,polarimetric2026,gaincorrection2025}. We therefore distinguish pulse extraction from BP prediction and do not interpret nonsignificant subgroup differences as parity.
\subsection{Evaluation Rigor and Data Leakage}
AuroraBP established calibration-based benchmarks and documented differences between laboratory and ambulatory performance \cite{aurorabp2022}. Record- or window-level splits can inflate reported accuracy when one participant contributes to training and testing \cite{bpbenchmark2023,mimicbp2024}, while dataset shortcuts may further obscure the information contributed by PPG \cite{limitations2024,challenges2024}. Our audit addresses a related change-prediction artifact: a calibration-row indicator may reveal observations whose target is near zero even when participant folds are disjoint.
\section{Framework and Audit Controls}
\label{sec:framework}
\subsection{Attractor Feature Bank}
A sampled PPG window is embedded as $\mathbf X_n=[x_n,x_{n+\tau},\ldots,x_{n+(d-1)\tau}]$, with $d=4$ and $\tau=5$ samples \cite{takens1981}. The eleven descriptors comprise waveform dispersion, skewness, kurtosis, sample and permutation entropy, recurrence rate, determinism, recurrence entropy, CSI, spectral entropy and a band-power quality ratio \cite{eckmann1987,marwan2007,richman2000,bandt2002}. Supplementary Section E gives filtering, sample limits, tolerances and the CSI equation. These are empirical descriptors, not validated physiological laws.

AuroraBP uses measured \texttt{raw\_*} waveforms and heuristic-corrected \texttt{pc\_*} channels. The auxiliary simulated stress-test \texttt{ref\_*} channel is not physiological ground truth and never enters the primary AuroraBP core.

\subsection{Pigmentation Correction Operator}
\label{sec:pcop}
The correction rescales the median-centered waveform using recorded Fitzpatrick type and blends it with Wiener filtering; it estimates neither melanin nor a Beer--Lambert parameter. Supplementary Section E specifies the complete operator and noise-dependent entropy/recurrence tolerances. Corrected-minus-raw MAE was $-0.001$~mmHg SBP (95\% CI $-0.037$--$0.034$) and $-0.031$~mmHg DBP ($-0.061$--$0.002$). The tested correction-and-extraction configuration has no established overall benefit or equivalence.

\subsection{Calibration Dose and Anchoring}
\label{sec:anchoring}
For participant $p$, the arithmetic mean $b_p^{(k)}$ of exactly $k\in\{1,2,3\}$ initial cuff readings defines
\begin{equation}
\Delta_{\mathrm{session}}^{(k)}(p,i)=y_{p,i}-b_p^{(k)}.
\label{eq:sessionanchor}
\end{equation}
The historical prior-anchored alternative uses the most recent valid cuff value $y_{p,i-1}^{*}$:
\begin{equation}
\Delta_{\mathrm{prior}}(p,i)=y_{p,i}-y_{p,i-1}^{*}.
\label{eq:prioranchor}
\end{equation}
Primary tracking uses the initial three-reading anchor without later cuff updates. Initial-visit targets occur up to 0.728~h after that anchor; return-visit targets occur 21.051--32.853~h after it. One-session calibration therefore does not imply a horizon shorter than one hour. Historical prior anchoring can use readings only minutes old within a visit, although across-visit gaps exceed one day; approximately 84\% of its reconstructed fixed-split gain came from anchor replacement. Neither scheme establishes cuff replacement. Calibration retention predicts $\widehat\Delta=0$.

\subsection{Context Operator}
\label{sec:context}
Wearable-state context comprises protocol-provided posture/activity and supplied optical quality. These states would require sensing or classification in deployment; they were not independently estimated here. The protocol-enriched comparator additionally uses phase, recording duration and pressure-quality metadata and is interpreted as an internal upper bound. A gradient-boosted tree model predicts BP change from the declared features and anchor. The original no-added-context arm shares a selected core that may contain elapsed-time terms, PPG descriptors and the UCI-derived PPG prior. It is therefore not a no-PPG condition. The matched factorial below supplies true optical-input-free controls.

\subsection{Calibration-Row Leakage Audit}
\label{sec:leakage}
A calibration indicator could identify trivial rows if a target reading were compared with itself, allowing a model to learn ``calibration row $\Rightarrow$ predict no change.'' We used four checks:
\begin{itemize}
\item[\textbf{A}] Compare $|\Delta|$ on flagged and unflagged rows.
\item[\textbf{B}] Re-score the same fitted models with flagged rows removed from the holdout only.
\item[\textbf{C}] Refit with the calibration-flag feature removed, flagged rows still scored.
\item[\textbf{D}] \textbf{(Decisive)} Remove the flag and exclude flagged rows from fitting and scoring.
\end{itemize}
Checks A--C each leave one channel open; Check D restricts both fitting and evaluation. The supplementary calibration-row leakage subsection reports all four.
\begin{finding}
On this dataset, calibration-flagged rows are \emph{not} near-zero-delta (mean $|\Delta|=6.774$ vs.\ $9.902$~mmHg SBP on development rows -- smaller, but not trivially so), and Check D confirms a positive context gain remains under strict exclusion ($+0.598$~mmHg SBP, 95\% CI $0.337$--$0.871$; $+0.557$~mmHg DBP, $0.315$--$0.795$). The separate historical attribution retains calibration-labelled rows and must not be interpreted as Check D attribution (see the supplementary calibration-row leakage analysis).
\end{finding}
\section{Methodology}
\label{sec:methods}
\subsection{Datasets}
\label{sec:datasets}
\textbf{AuroraBP} is the primary dataset \cite{aurorabp2022,aurorabpDataset}. It contains 1{,}221 participant records and 11{,}597 auscultatory measurements across calibration, static, exercise, and temporal-challenge phases; waveforms were resolvable for 672 participants. The primary common cohort includes 653 participants with three initial calibration readings. Excluding 1{,}337 later calibration-labelled rows leaves 7{,}879 scored measurements. Historical ablations used a fixed 501/168 split; because its holdout was inspected during development, it is treated as historical evidence rather than untouched validation.
Four secondary sources appear in the development notebooks: BIDMC, MIMIC-II, Welltory-PPG, and UCI \cite{bidmcDataset,pimentel2017,physionet2000,mimicii2011,mimiciiDataset,welltory2021,uciCufflessDataset,kachuee2015}. The first three form a historical corrected-feature PCA bank whose coordinates are excluded from the primary candidate list. UCI trains the BP-change prior that may enter the primary model. Pigmentation ranks on these sources are synthetic simulation inputs, not measured traits, and support no equity claim.

Two datasets support independent follow-up analyses. The \textbf{Appel--Theart Participant Study Data} provide paired visible and infrared PPG, Monk Skin Tone (MST), and pulse-rate references \cite{appel2025ppg}. After auditing, 108 participants contributed 630 pulse-rate windows, including 24 participants with MST~7--10. Its single cuff BP value per participant permits only cross-sectional BP analysis. \textbf{VitalDB} provides synchronized intraoperative PPG and arterial BP \cite{lee2022vitaldb,vitaldbDataset}. A locked evaluation used 5{,}472 future windows from 228 cases belonging to 227 subjects disjoint from the 55 development subjects. Cases were grouped by subject for inference. VitalDB has no skin-tone field and tests a separately developed intraoperative implementation, not transfer of AuroraBP-fitted parameters.
Table~\ref{tab:cohortflow} consolidates participant eligibility and the relationship between the historical and primary cohorts.\begin{table}[t]
\centering
\caption{AuroraBP participant flow and analysis populations. Release-to-waveform counts were checked against the original participant and measurement tables; downstream eligibility and cohort overlap were reconciled using the cached feature-bank participant identifiers.}
\label{tab:cohortflow}
\small
\begin{tabular}{p{4.5cm}rp{7.3cm}}
\toprule
Stage & Participants & Eligibility or transition \\
\midrule
Released dataset & 1,221 & 673 participants have 11,597 auscultatory measurements; 548 released participants have no auscultatory records. \\
Cached waveform features & 672 & 11,427 extracted rows from 672 participants. One of the 673 auscultatory participants has no available waveform; together with the 548 above, this accounts for the 549 outside this bank. \\
Historical calibration cohort & 669 & At least two valid initial calibration readings and future measurements. Three participants with only one initial reading are excluded. The historical anchor averages available initial calibration replicates. \\
Historical development/evaluation & 501 / 168 & Fixed split within the preceding 669. Evaluation participants were inspected across development analyses. \\
Primary common cohort & 653 & At least three valid initial calibration readings; 19 of the 672 have insufficient initial readings (16 have two; three have one). All 653 also belong to the historical cohort; 489/164 were in its development/evaluation groups. \\
Primary measurement endpoint & 653 & 9,216 eligible future rows before excluding 1,337 later calibration-labelled rows; 7,879 readings remain. Initial anchor readings do not enter this future-row count or the fitted/scored endpoint. \\
\bottomrule
\end{tabular}
\end{table}

Table~\ref{tab:datasetsummary} consolidates the analysis sample, role, and source references for every dataset used in the study.
\begin{table*}[t]
\centering
\caption{Datasets and analysis roles. AuroraBP access requires provider approval and a Data Use Agreement. Access details are consolidated in Data Availability. UCI-derived inputs are excluded from the clean factorial.}
\label{tab:datasetsummary}
\scriptsize
\renewcommand{\arraystretch}{1.12}
\begin{tabular}{>{\raggedright\arraybackslash}p{1.7cm}>{\raggedright\arraybackslash}p{3.0cm}>{\raggedright\arraybackslash}p{4.4cm}>{\raggedright\arraybackslash}p{5.2cm}}
\toprule
Dataset & Material analyzed & Role & Source \\
\midrule
AuroraBP & 653 participants; 7,879 scored measurements & Primary calibrated BP evaluation & Microsoft Research / Zenodo \cite{aurorabp2022,aurorabpDataset} \\
BIDMC & 53 subjects; 1,272 windows & Historical PCA representation only & PhysioNet \cite{bidmcDataset,pimentel2017} \\
MIMIC-II & 1,500 sampled windows & Historical PCA representation only & PhysioNet \cite{mimicii2011,mimiciiDataset} \\
Welltory-PPG & 106 windows & Historical PCA representation only & Published PPG collection \cite{welltory2021} \\
UCI Cuff-Less BP & 425 records; 3,268 windows & Prior training for original comparison & UCI repository \cite{uciCufflessDataset,kachuee2015} \\
Appel--Theart & 108 participants; 630 windows & Independent pulse-rate fusion evaluation & Mendeley Data \cite{appel2025ppg} \\
VitalDB & 55 development / 227 holdout subjects; 5,472 test windows & Separate calibrated BP evaluation & PhysioNet \cite{lee2022vitaldb,vitaldbDataset} \\
\bottomrule
\end{tabular}
\begin{minipage}{0.96\textwidth}
\vspace{2pt}\footnotesize Counts describe the material used by this study and need not equal the full source repository. Auxiliary-source pigmentation ranks were synthetic simulation inputs; only AuroraBP supplied Fitzpatrick labels and Appel--Theart supplied Monk Skin Tone labels.
\end{minipage}
\end{table*}
\subsection{Feature Extraction and Selection}
\label{sec:features}
Standard AuroraBP descriptors use the central 10-s optical window, third-order 0.5--8~Hz bandpass and timestamp-derived 500~Hz sampling. All 11,427 feature rows matched the timestamp sampling audit within $2\times10^{-11}$~Hz. Derived \texttt{raw\_ppg\_quality} is a Welch band-power ratio (0.5--5~Hz divided by 0.1--15~Hz), distinct from the supplied \texttt{optical\_quality} context annotation. CSI recomputation agreed within $2.3\times10^{-16}$. Full extraction specifications are in Supplementary Section E.

Within each outer training set, the original primary procedure ridge-residualizes against baseline BP/log elapsed time, ranks eligible raw-feature changes, relative changes, time terms and the UCI prior using mutual information, and retains up to 24 candidates plus baseline BP. Context arms share that selected core. The UCI prior was retained in 26/30 endpoint/fold cores and uses corrected optical features, so excluding explicit Fitzpatrick terms cannot establish absence of indirect Fitzpatrick dependence. Historical PCA coordinates and \texttt{ref\_*} features are excluded. Historical HCA augmentation is distinct from the matched nonlinear/attractor family.
\subsection{Model and Training}
The primary model uses training-fitted median imputation and squared-loss HistGradientBoosting: learning rate 0.03, 15 leaves, regularization 1.0, 300 iterations and seed 42. Participant weights are inverse row counts, clipped at training-set 2nd/98th percentiles and normalized. Settings were inherited from development. The change gate uses fitted training predictions rather than independent inner validation; activation was 1.0. This remains a development limitation.
\subsection{Evaluation Protocol}
\label{sec:evalprotocol}
The common cohort contains 653 participants and 7,879 future non-calibration readings. Three repeats of five participant folds, stratified by Fitzpatrick band with seeds $42+26200+r$ for zero-based repeat $r$, repeat ranking, imputation, fitting and gate selection using outer training participants only. Initial anchor readings and later calibration-labelled targets are excluded from fitting/scoring. Within each repeat, MAE is averaged within participant and then equally across participants; participant MAEs are subsequently averaged across repeats. Bootstrap inference resamples participants with their readings/repeats together. AuroraBP informed earlier development, so this is retrospective internal validation.

Historical 501/168 analyses are reconstructed development evidence, not untouched validation. Retained UCI-prior defaults were used; re-executing the recovered tuning procedure selected those same settings under the current inputs/runtime, without recovering the original Colab state or tuning CSV. Historical intervals use 4,000 participant resamples. Supplementary Section E documents this distinction and the remaining reconstruction details.
\subsubsection{Claim hierarchy and multiplicity}
For participant $p$, repeat $r$ and arm $a$, let $L_{pra}=n_p^{-1}\sum_i|\widehat y_{pira}-y_{pi}|$. Primary gain is
\begin{equation}
G=\frac1P\sum_p\left[\frac1R\sum_r L_{pr,\mathrm{no\ context}}-\frac1R\sum_r L_{pr,\mathrm{wearable\ state}}\right].
\label{eq:primarygain}
\end{equation}
Positive gain is an absolute MAE reduction in mmHg: rounded examples are $8.133-7.697=0.436$ SBP and $6.265-6.139=0.126$ DBP. Intervals use unrounded pairs and 10,000 participant resamples. Individual primary gain intervals are 97.5\%, targeting approximate Bonferroni simultaneous 95\% coverage for SBP/DBP. Holm adjustment applies separately to the two paired sign-flip tests (99,999 assignments; seed 20260910); paired $t$ tests are sensitivity analyses. Sign exchangeability and participant sampling assumptions apply. Neither bootstrap nor multiplicity correction removes dependence from shared training sets or retrospective development. Arm estimates and historical/reviewer secondary contrasts use marginal 95\% intervals. Later follow-up interval levels are specified separately below.

The matched feature-family ablation shares calibration, time, wearable-state context, folds, fixed estimator, weights and training-stage gate, excluding UCI. Standard descriptors are dispersion, skewness, kurtosis, spectral entropy and the band-power ratio; the augmentation adds sample/permutation entropy, recurrence measures and CSI. Pulse-arrival time was entirely missing and omitted. Fixed semantic sets replace feature ranking. This tests the combined nonlinear/attractor addition; its two endpoint sign-flip tests receive a separate Holm adjustment.
\subsubsection{Matched context-by-PPG and time-removal analyses}
\label{sec:factorialmethods}
Four reviewer-requested arms share endpoint baseline BP and the same cohort/folds: A, calibration/time; B, A plus protocol posture/activity; C, A plus PPG; D, A plus both. No factorial uses UCI, corrected features, PCA, PAT, phase, duration, pressure quality or measurement order. Supplied optical quality belongs to the PPG block alongside changes/relative changes in all eleven raw optical descriptors, making A/B genuinely optical-input-free. All arms use the same fixed estimator, training-only imputation/weights and original gate procedure, with no new tuning. Constant/missing columns are removed using training data only.

A matched sensitivity removes elapsed seconds, log elapsed days, session elapsed seconds and log session elapsed days. Additional arms omit either posture or activity. PPG gains are $M_A-M_C$ and $M_B-M_D$; context gains are $M_A-M_B$ and $M_C-M_D$. Interaction is $(M_C-M_D)-(M_A-M_B)$ on the MAE scale, not physiological synergy. Paired intervals use 10,000 participant resamples (seed 20261019), are exploratory marginal 95\%, and do not extend the primary test family.
\subsubsection{Signed errors and repeated-measures agreement}
\label{sec:agreementmethods}
Unchanged primary out-of-fold predictions supply prediction-minus-reference bias, individual-error SD and absolute-error tail fractions. Moments are averaged within participant, equally across participants, and separately per repeat; SD is the square root of mean squared error minus squared bias. Repeat predictions are not pooled as independent observations or averaged into an unreported ensemble.

Approximate agreement limits use separately estimated within-/between-participant random-intercept variance components \cite{bland2007repeated}, centered on participant-balanced bias. Supplementary Section E gives the full method-of-moments equations and this adaptation's distinction from the cited weighted-mean estimator. Statistics and limits are calculated per repeat and averaged; participant bootstrap preserves repeats together. Homogeneous variation and conditionally independent residuals are approximations: serial dependence, nonnormal tails and BP-dependent errors can limit coverage. These limits are descriptive, not device certification.
\subsubsection{Additional evaluation protocols}
Historical pigmentation models are refitted on 501 development participants and evaluated on the reused 168-participant group; primary subgroup summaries use all 653 participants. Small V--VI groups (6/25) preclude fairness conclusions. The nested endpoint-specialization/router analyses are exploratory; the DBP router only requests repeat measurement and never changes the prediction (Supplementary Sections B and E).

The independent pulse analysis uses visible/infrared quality-weighted pulse estimates and an MST-dependent guard selected in four participant inner folds. Twenty repeats of five outer folds average repeated predictions per window before participant MAE: these are repeated-cross-fit ensemble estimates. Equal weighting is a fixed post-hoc comparator on the same 630 windows; paired intervals use 108 participants and 10,000 resamples. Supplementary Section A specifies processing and guard activation.

VitalDB uses median BP/features from three early arterial-reference windows per case, excluding them from fitting/scoring. Context comprises anchor, time, age, sex and hypertension; PPG adds current/relative signal descriptors. Development-only selection in 55 subjects froze a constant 50/50 context/PPG blend before evaluation on 228 cases from 227 disjoint subjects and 5,472 future windows. Bootstrap groups cases by subject. This is a separate intraoperative implementation, not AuroraBP-model transfer. Reference/quality exclusions do not use prediction error; neither error-based removal nor holdout retuning was performed. Supplementary Sections A/E give selection, clipping and alignment details.
\subsection{Later Exploratory Signal and Sensor-History Follow-ups}
\label{sec:followupmethods}
A separate series retains the same 653 participants, 7,879 records, PPG/accelerometer modalities and three initial cuff readings. Its static-rich reference adds current, difference and relative PPG/beat/accelerometer descriptors, initial SBP/DBP means/SDs and elapsed time. Missing descriptors are training-imputed without dropping records. Explicit ECG/PAT, protocol annotations, pressure quality, Fitzpatrick predictors, corrected optical variables and UCI are excluded. These features/estimators differ from the original primary model.

Three repeats of five participant folds use seeds $42+1000r$, different from the primary seeds but matched within this series. Three-fold participant inner predictions select the static-rich blend of absolute-/squared-loss boosting, extra trees and retention. Subsequent fixed blends use $0.25/0.25/0.50$ weights for the three estimators. Follow-ups test enriched median/calibration candidates, direct BP targets and 72 recording-level history features from 18 PPG/accelerometer descriptors. History uses only earlier sensor records, resets after gaps over one hour and never supplies later cuff values; its initial sensor anchor is recovered from the permitted calibration summary where available.

Each specification was frozen before its own outer predictions, after inspection of earlier experiments. Gains use paired participant MAEs and 5,000 bootstrap resamples. Static-rich-reference comparisons retain individual 97.5\% endpoint intervals; matched fixed-weight contrasts use 95\%. Neither corrects for successive adaptive experiments. Supplementary Sections D/E contain full specifications and reproduction boundaries.

\section{Experiments and Results}
\label{sec:results}
\subsection{Primary Repeated Participant-Level Validation}
\label{sec:v26primary}
Table~\ref{tab:v26primary} reports the original comparison on 653 participants and 7,879 strict-endpoint measurements. Wearable-state MAEs were $7.697$/$6.139$~mmHg SBP/DBP, with paired gains $0.436$/$0.126$ and 97.5\% intervals $[0.305,0.573]$/$[0.053,0.200]$. Holm-adjusted sign-flip $p$ values were $0.00002$/$0.00013$; paired $t$ sensitivity tests also passed. Optical quality alone lowered SBP error by $0.022$~mmHg but increased DBP error by $0.008$, offering no meaningful incremental benefit. The primary band means in Supplementary Table S8, weighted by counts 516/112/25, reproduce the overall estimates before rounding. Posture/activity remain protocol annotations; the protocol-enriched arm is an internal upper bound.

\begin{table}[t]
\centering
\caption{Original repeated participant-level validation (653 participants; three-reading anchor). Arm intervals are marginal 95\%; the two gain intervals are 97.5\% to target approximate 95\% simultaneous coverage by a Bonferroni rationale. Holm adjustment applies separately to paired test $p$ values. The shared core can contain PPG descriptors and the UCI-derived prior; ``no added context'' is not a no-PPG control.}
\label{tab:v26primary}
\scriptsize
\resizebox{\columnwidth}{!}{%
\begin{tabular}{lrr}
\toprule
Arm & SBP MAE [95\% CI] & DBP MAE [95\% CI] \\
\midrule
Calibration retention & 9.790 & 6.767 \\
No added context & 8.133 [7.824, 8.455] & 6.265 [6.019, 6.526] \\
Optical quality only & 8.110 [7.802, 8.431] & 6.273 [6.025, 6.535] \\
Wearable-state context & 7.697 [7.407, 8.008] & 6.139 [5.899, 6.391] \\
Protocol-enriched upper bound & 7.523 [7.243, 7.815] & 5.743 [5.527, 5.977] \\
\midrule
Primary gain [97.5\% CI] & 0.436 [0.305, 0.573] & 0.126 [0.053, 0.200] \\
Holm-adjusted paired $p$ & 0.00002 & 0.00013 \\
\bottomrule
\end{tabular}}
\end{table}
\subsection{Matched Context-by-PPG Factorial and Time Sensitivity}
\label{sec:factorial}
Table~\ref{tab:factorial} separates the information supplied by calibration/time, protocol-derived posture/activity context, and PPG without using the UCI prior. With elapsed-time terms, PPG reduced MAE by $0.315$~mmHg SBP (95\% CI $0.175$--$0.456$) and $0.209$~mmHg DBP ($0.103$--$0.316$) without context, and by $0.214$~mmHg SBP ($0.111$--$0.319$) and $0.135$~mmHg DBP ($0.046$--$0.225$) with context. Context gains were $0.503$/$0.209$~mmHg without PPG and $0.402$/$0.135$~mmHg with PPG. These contrasts are conditional on the declared block definitions and fixed estimator, not measures of physiological causation. Supplementary Table S10 reports all conditional and interaction contrasts.

Without elapsed-time terms, calibration-only MAE was 10.021~mmHg for SBP and 6.807~mmHg for DBP. The context-plus-PPG arm achieved 7.771 and 6.183~mmHg, respectively. PPG gains with context were $0.305$~mmHg SBP (95\% CI $0.187$--$0.429$) and $0.087$~mmHg DBP ($-0.009$--$0.181$); the DBP interval includes zero. Context gains with PPG remained $1.198$/$0.299$~mmHg. Thus clock-derived covariates account for substantial predictive information, and the small DBP PPG increment is sensitive to their inclusion. Persistence of annotation gains without time does not establish real-world transportability.

The leave-one-annotation-out analysis (Supplementary Table S11) further shows that, without clock terms, adding the activity label to a PPG-plus-posture model reduces MAE by $0.823$~mmHg SBP ($0.685$--$0.958$) and $0.245$~mmHg DBP ($0.161$--$0.329$). Adding posture to PPG-plus-activity gives $0.093$~mmHg SBP ($0.060$--$0.125$) and $-0.004$~mmHg DBP ($-0.012$--$0.004$). The time-free context result therefore depends particularly on the protocol-provided activity annotation. No independently estimated state classifier has been evaluated.

\begin{table}[t]
\centering
\caption{Reviewer-requested matched factorial, with and without time terms. All arms use the same 653 participants, folds, fixed estimator, weights and training-stage gate; no arm uses the UCI prior. A/B contain no optical inputs. Context is protocol-provided posture/activity; supplied optical quality is assigned to the PPG block. Intervals are exploratory marginal 95\% participant-bootstrap intervals.}
\label{tab:factorial}
\small
\begin{tabular}{lrr}
\toprule
Arm & SBP MAE [95\% CI] & DBP MAE [95\% CI] \\
\midrule
\multicolumn{3}{l}{\textit{With elapsed-time terms}} \\
A: Calibration & 8.426 [8.098, 8.768] & 6.472 [6.215, 6.751] \\
B: Calibration + context & 7.923 [7.610, 8.243] & 6.263 [6.015, 6.532] \\
C: Calibration + PPG & 8.111 [7.806, 8.432] & 6.263 [6.015, 6.525] \\
D: Calibration + context + PPG & 7.709 [7.414, 8.015] & 6.128 [5.888, 6.383] \\
\midrule
\multicolumn{3}{l}{\textit{Without elapsed-time terms}} \\
A: Calibration & 10.021 [9.673, 10.383] & 6.807 [6.530, 7.101] \\
B: Calibration + context & 8.076 [7.749, 8.411] & 6.269 [6.016, 6.544] \\
C: Calibration + PPG & 8.969 [8.644, 9.298] & 6.481 [6.223, 6.749] \\
D: Calibration + context + PPG & 7.771 [7.478, 8.077] & 6.183 [5.945, 6.436] \\
\bottomrule
\end{tabular}
\end{table}
\subsection{Primary Signed Errors and Agreement}
\label{sec:agreement}
\begin{figure}[t]
\centering
\includegraphics[width=\linewidth]{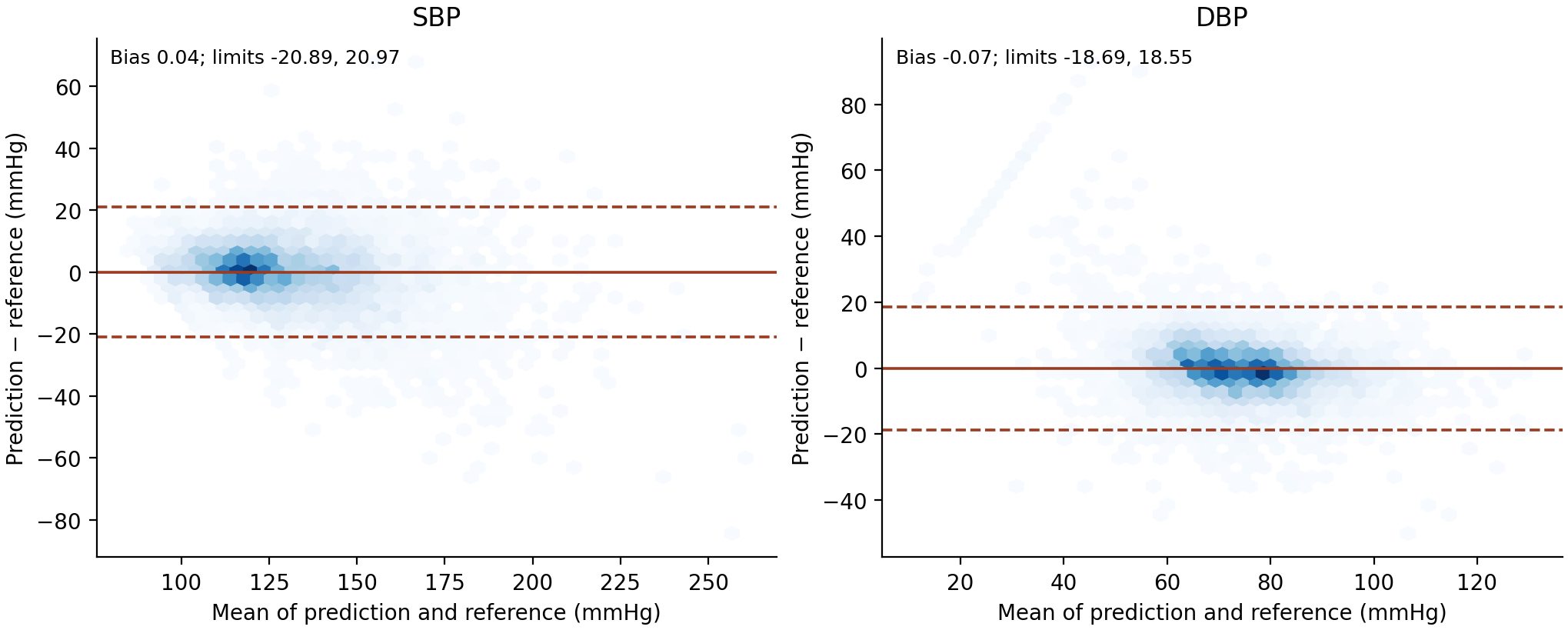}
\caption{Agreement for the original wearable-state model in the first cross-validation repeat. Hexagonal bins show individual reading density, with prediction minus reference on the vertical axis. Solid lines mark participant-balanced bias; dashed lines mark approximate repeated-measures limits using within- and between-participant error variances. Limits describe individual-reading agreement under variance-model assumptions, not clinical interchangeability.}
\label{fig:primaryagreement}
\end{figure}

Table~\ref{tab:agreement} reports signed bias and individual-reading dispersion from the unchanged original predictions. The wearable-state arm has mean bias $0.038$~mmHg SBP (95\% CI $-0.501$--$0.561$) and $-0.091$~mmHg DBP ($-0.482$--$0.309$), but participant-balanced error SDs remain $10.957$ and $9.641$~mmHg. Absolute errors exceed 15~mmHg for $13.35\%$ and $7.03\%$ of readings under equal participant weighting. Approximate repeated-measures limits of agreement, averaged over the three separately assessed repeats, are $[-21.084,21.159]$~mmHg SBP and $[-18.756,18.574]$~mmHg DBP. Near-zero mean bias therefore coexists with substantial individual errors. The error cloud is heterogeneous across BP levels, so the approximate limits should not be assumed to provide uniform 95\% coverage throughout the BP range. Figure~\ref{fig:primaryagreement} displays the first repeat; Supplementary Table S13 reports uncertainty for dispersion and agreement limits.

\begin{table}[t]
\centering
\caption{Signed errors in the original primary comparison. Bias is prediction minus reference; positive denotes overestimation. SD is participant-balanced individual-error dispersion. Statistics are calculated separately per repeat and then averaged; MAEs are reported in Table~\ref{tab:v26primary}.}
\label{tab:agreement}
\small
\begin{tabular}{llrrr}
\toprule
Endpoint & Arm & Bias (mmHg) & Error SD (mmHg) & $|e|>15$~mmHg (\%) \\
\midrule
SBP & No added context & -0.009 & 11.605 & 14.75 \\
SBP & Wearable-state context & 0.038 & 10.957 & 13.35 \\
DBP & No added context & -0.086 & 9.846 & 7.19 \\
DBP & Wearable-state context & -0.091 & 9.641 & 7.03 \\
\bottomrule
\end{tabular}
\end{table}

\subsection{Matched Feature-Family Ablation}
\label{sec:featureablation}
Table~\ref{tab:featureablation} reports the secondary matched ablation. Adding the standard PPG descriptors to calibration, time, and wearable-state context reduced participant MAE by $0.187$~mmHg SBP (95\% CI $0.088$--$0.286$) and $0.180$~mmHg DBP ($0.096$--$0.266$). Adding the nonlinear/attractor family to that standard arm produced only $0.025$~mmHg SBP ($-0.018$ to $0.069$) and $0.009$~mmHg DBP ($-0.033$ to $0.050$) further reductions. The Holm-adjusted paired sign-flip $p$ values for those latter contrasts were $0.532$ and $0.662$. Thus the matched analysis supports an incremental contribution from the declared standard PPG descriptors but does not establish an attractor-family advantage.
\begin{table}[t]
\centering
\caption{Secondary Matched AuroraBP Feature-Family Ablation (653 Participants)}
\label{tab:featureablation}
\scriptsize
\resizebox{\columnwidth}{!}{%
\begin{tabular}{lrr}
\toprule
Arm & SBP MAE [95\% CI] & DBP MAE [95\% CI] \\
\midrule
Calibration, time, wearable-state context & 7.920 [7.607, 8.256] & 6.318 [6.067, 6.589] \\
+ Standard PPG descriptors & 7.733 [7.441, 8.048] & 6.137 [5.898, 6.388] \\
+ Nonlinear/attractor family & 7.709 [7.418, 8.017] & 6.128 [5.887, 6.384] \\
\midrule
Attractor-family gain [95\% CI] & 0.025 [$-0.018$, 0.069] & 0.009 [$-0.033$, 0.050] \\
Holm-adjusted paired $p$ & 0.532 & 0.662 \\
\bottomrule
\end{tabular}}
\end{table}
\subsection{Calibration Dose}
\label{sec:caldose}
Table~\ref{tab:caldose} reports protocol-enriched models fitted on identical participants and future measurements for each calibration dose. Ranking and fitting are repeated within the same participant folds. These dose gains belong to the protocol-enriched arm, not the lead wearable-state arm. Three readings are one-session repeated calibration, not one-reading calibration.
\begin{table}[t]
\centering
\caption{Refitted Protocol-Enriched Models by Calibration Dose (653 Participants)}
\label{tab:caldose}
\scriptsize
\resizebox{\columnwidth}{!}{%
\begin{tabular}{crrrr}
\toprule
Target & Readings & Retention & Model MAE [95\% CI] & Gain vs. retention \\
\midrule
SBP & 1 & 10.547 & 8.257 [7.933, 8.602] & 2.290 [2.051, 2.525] \\
SBP & 2 & 10.037 & 7.750 [7.458, 8.055] & 2.286 [2.066, 2.509] \\
SBP & 3 & 9.790 & 7.523 [7.243, 7.815] & 2.267 [2.044, 2.489] \\
DBP & 1 & 7.213 & 6.280 [6.026, 6.548] & 0.934 [0.739, 1.129] \\
DBP & 2 & 6.900 & 5.873 [5.648, 6.112] & 1.027 [0.833, 1.215] \\
DBP & 3 & 6.767 & 5.743 [5.527, 5.977] & 1.024 [0.832, 1.211] \\
\bottomrule
\end{tabular}}
\end{table}
\subsection{Decomposition of Historical Prior-Anchor Improvement}
\label{sec:v21}
Table~\ref{tab:v21} reports the reconstructed fixed-feature ladder. The net SBP reduction is $2.069$~mmHg ($26.0\%$), of which approximately $84\%$ is attributable to replacing the initial anchor with the most recent prior cuff reading. The reference can be minutes old within a visit or more than one day old across visits. This comparison quantifies dependence on the reference scheme, not cuff replacement. The historical HCA addition worsens SBP MAE by $0.128$~mmHg (95\% CI $0.019$--$0.242$); this finding is distinct from the matched primary feature-family test.
\begin{table}[t]
\centering
\caption{Reconstructed historical isolation ladder (SBP; common development-selected feature set; 168 evaluation participants). The fixed evaluation group was inspected during development; intervals are marginal and do not represent untouched independent validation.}
\label{tab:v21}
\small
\begin{tabular}{lrrr}
\toprule
Step & MAE & Gain [95\% CI] & Share \\
\midrule
Session anchor (legacy) & 7.970 & -- & -- \\
$\rightarrow$ Prior anchor & 6.223 & $1.747$ [1.298, 2.200] & 84.4\% \\
$\rightarrow$ Absolute-error loss & 6.110 & $0.113$ [-0.013, 0.249] & 5.5\% \\
$\rightarrow$ +Context & 5.773 & $0.337$ [0.209, 0.467] & 16.3\% \\
$\rightarrow$ +HCA features & 5.901 & $-0.128$ [-0.242, -0.019] & -6.2\% \\
\midrule
Total (session $\rightarrow$ full) & & $2.069$ [1.617, 2.516] & 100\% \\
\bottomrule
\end{tabular}
\par\smallskip\footnotesize Gain is preceding minus subsequent MAE; negative values indicate worsening. Shares divide each step by the net total gain.
\end{table}

\subsection{Secondary Pigmentation-Intervention Audit}
\label{sec:equity}
Table~\ref{tab:equity} compares reconstructed configurations on the prior-anchored absolute-error-plus-context base. The reference has no added Fitzpatrick terms, although indirect dependence through the corrected-feature UCI prior is possible. Continuous Fitzpatrick terms/interactions, band reweighting and group gates are tested as additions. All configurations share the same participants, targets, base features and estimator. Reweighting and group gating are not isolated from one another.
\begin{table}[t]
\centering
\caption{Reconstructed historical Fitzpatrick-input comparison (168 evaluation participants). The fixed evaluation group was reused during development; this is exploratory internal evidence.}
\label{tab:equity}
\small
\begin{tabular}{p{8cm}rr}
\toprule
Configuration & SBP MAE & DBP MAE \\
\midrule
Reference & 5.773 & 4.483 \\
Fitzpatrick inputs only & 5.770 & 4.479 \\
+ Band reweighting and group gating & 6.073 & 4.710 \\
Full Fitzpatrick-input configuration & 6.073 & 4.710 \\
\midrule
\multicolumn{3}{l}{Reference minus full MAE [95\% CI]; negative favors reference:} \\
SBP & \multicolumn{2}{l}{$-0.300$ [-0.419, -0.182]} \\
DBP & \multicolumn{2}{l}{$-0.227$ [-0.355, -0.114]} \\
\bottomrule
\end{tabular}
\end{table}
Fitzpatrick terms alone give negligible point changes. The full configuration worsens SBP/DBP MAE by $0.300$/$0.227$~mmHg (95\% CIs $0.182$--$0.419$/$0.114$--$0.355$). Its between-band ranges also increase; six V--VI holdout participants cannot establish fairness or parity. Supplementary Sections B/E identify estimator-specific subgroup results and retain detailed diagnostics.
\subsubsection{Repeated-CV pigmentation-correction intervention}
Table~\ref{tab:pccorrection} reconstructs the separate historical strict-cohort channel ablation. The corrected channel alone has no established overall benefit; equivalence was not tested. The exploratory V--VI SBP contrast shows $0.206$~mmHg worsening (95\% CI $0.023$--$0.386$), whereas the I--II DBP contrast shows a small decrease. These marginal subgroup intervals have no multiplicity adjustment.
\begin{table}[t]
\centering
\caption{Reconstructed corrected-versus-raw channel ablation in repeated participant CV. This separate historical configuration is not the primary model hierarchy. Intervals are marginal 95\%.}
\label{tab:pccorrection}
\small
\begin{tabular}{llrrrr}
\toprule
Target & Scope & $n$ & Raw & Corrected & $\Delta$ MAE [95\% CI] \\
\midrule
SBP & Overall & 653 & 7.524 & 7.523 & $-0.001$ [-0.037, 0.034] \\
SBP & I--II & 516 & 7.487 & 7.484 & $-0.003$ [-0.040, 0.035] \\
SBP & III--IV & 112 & 7.683 & 7.645 & $-0.038$ [-0.143, 0.062] \\
SBP & V--VI & 25 & 7.571 & 7.777 & $0.206$ [0.023, 0.386] \\
DBP & Overall & 653 & 5.745 & 5.714 & $-0.031$ [-0.061, 0.002] \\
DBP & I--II & 516 & 5.683 & 5.642 & $-0.041$ [-0.077, -0.004] \\
DBP & III--IV & 112 & 6.065 & 6.072 & $0.006$ [-0.072, 0.096] \\
DBP & V--VI & 25 & 5.576 & 5.598 & $0.022$ [-0.082, 0.125] \\
\bottomrule
\end{tabular}
\par\smallskip\footnotesize $\Delta=$ corrected minus raw; negative favors correction.
\end{table}
The full corrected-channel-plus-Fitzpatrick configuration changes SBP MAE by $+0.003$~mmHg (95\% CI $-0.038$ to $0.043$) relative to raw. Its DBP decrease is $0.0382$~mmHg, with a marginal interval of $0.00004$--$0.07642$ before rounding. The latter narrowly excludes zero but is an exploratory, unadjusted contrast across several configurations; it does not establish a general pigmentation-correction benefit or equitable performance.

Participant-median signal diagnostics found Fitzpatrick-associated quality and dispersion differences, without identifying measured-melanin effects. Device gain, contact pressure, anatomy and protocol may confound these associations; coefficients and intervals are retained in Supplementary Section E.

The historical context comparison contains only six V--VI participants; its band-specific point changes cannot establish fairness (Supplementary Section E).

\subsection{Independent Quality-Fusion and Guarded-Correction Test}
\begin{figure}[t]
\centering
\includegraphics[width=\linewidth]{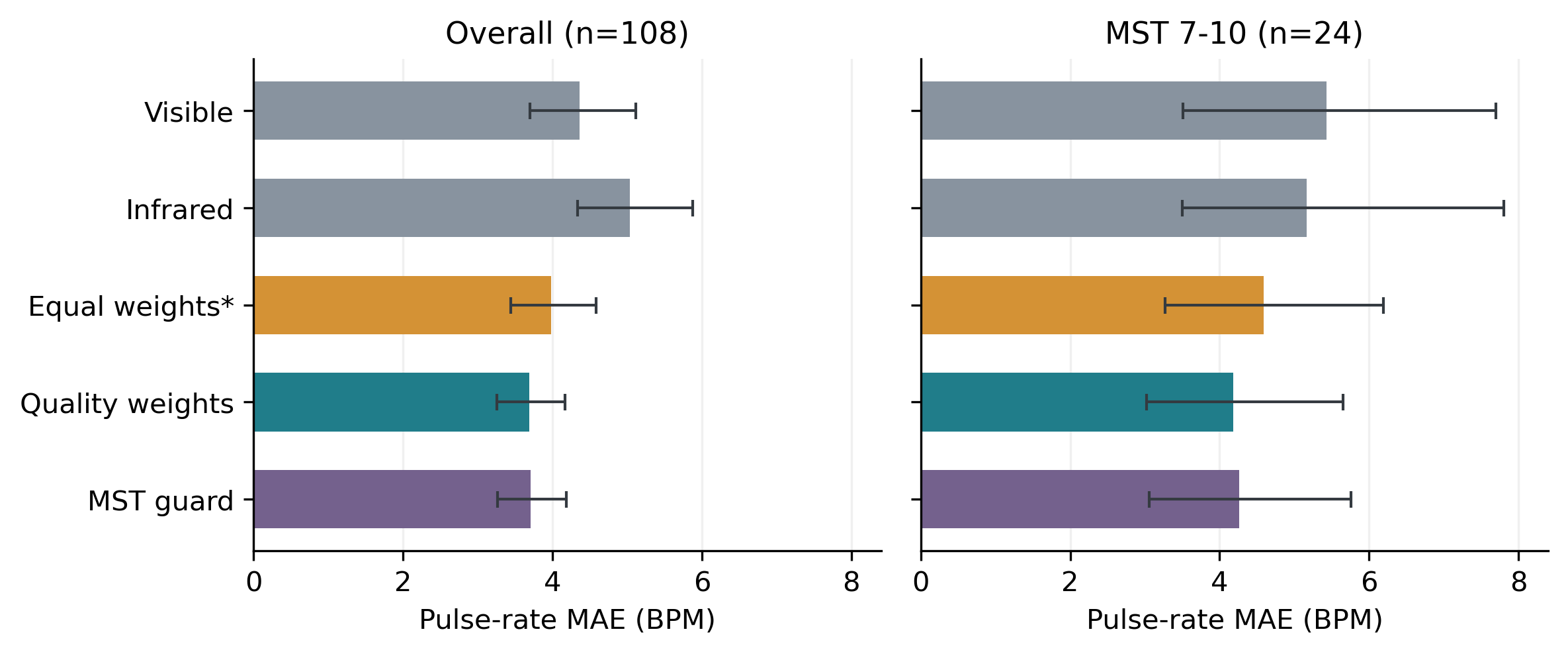}
\caption{Participant-balanced pulse-rate MAE with marginal 95\% bootstrap intervals. The equal-weight comparison was added post hoc; quality weighting improves overall error relative to equal weighting, while guarded MST correction does not establish additional benefit.}
\label{fig:mendeleyfusion}
\end{figure}

Table~\ref{tab:mendeleyfusion} and Figure~\ref{fig:mendeleyfusion} report the independent pulse-rate analysis. Quality weighting reduced MAE from $3.985$ to $3.690$~BPM versus the post-hoc equal-weight comparator: paired difference $-0.294$~BPM (95\% CI $-0.567$--$-0.069$). It also improved on either single wavelength. The MST~7--10 contrast versus equal weighting was $-0.402$~BPM ($-0.929$--$0.010$), without established subgroup benefit. The MST guard activated in 5\% of outer folds; its corrected-minus-tone-blind difference was $0.013$~BPM ($-0.010$--$0.053$). These marginal comparisons establish neither additional correction benefit nor BP utility. Supplementary Section E retains all pairwise estimates.
\begin{table}[t]
\centering
\caption{Independent Multi-Wavelength PPG Mechanism Test}
\label{tab:mendeleyfusion}
\scriptsize
\resizebox{\columnwidth}{!}{%
\begin{tabular}{lrr}
\toprule
Arm & Overall MAE (BPM) & MST~7--10 MAE \\
\midrule
Visible only & 4.366 & 5.434 \\
Infrared only & 5.037 & 5.165 \\
Equal-weight fusion (post hoc) & 3.985 & 4.587 \\
Quality-weighted fusion & \textbf{3.690} & \textbf{4.184} \\
Guarded MST-aware fusion & 3.703 & 4.265 \\
\bottomrule
\end{tabular}
}
\end{table}

\subsection{VitalDB Context-Plus-PPG Frozen Holdout}
\begin{figure}[t]
\centering
\includegraphics[width=\linewidth]{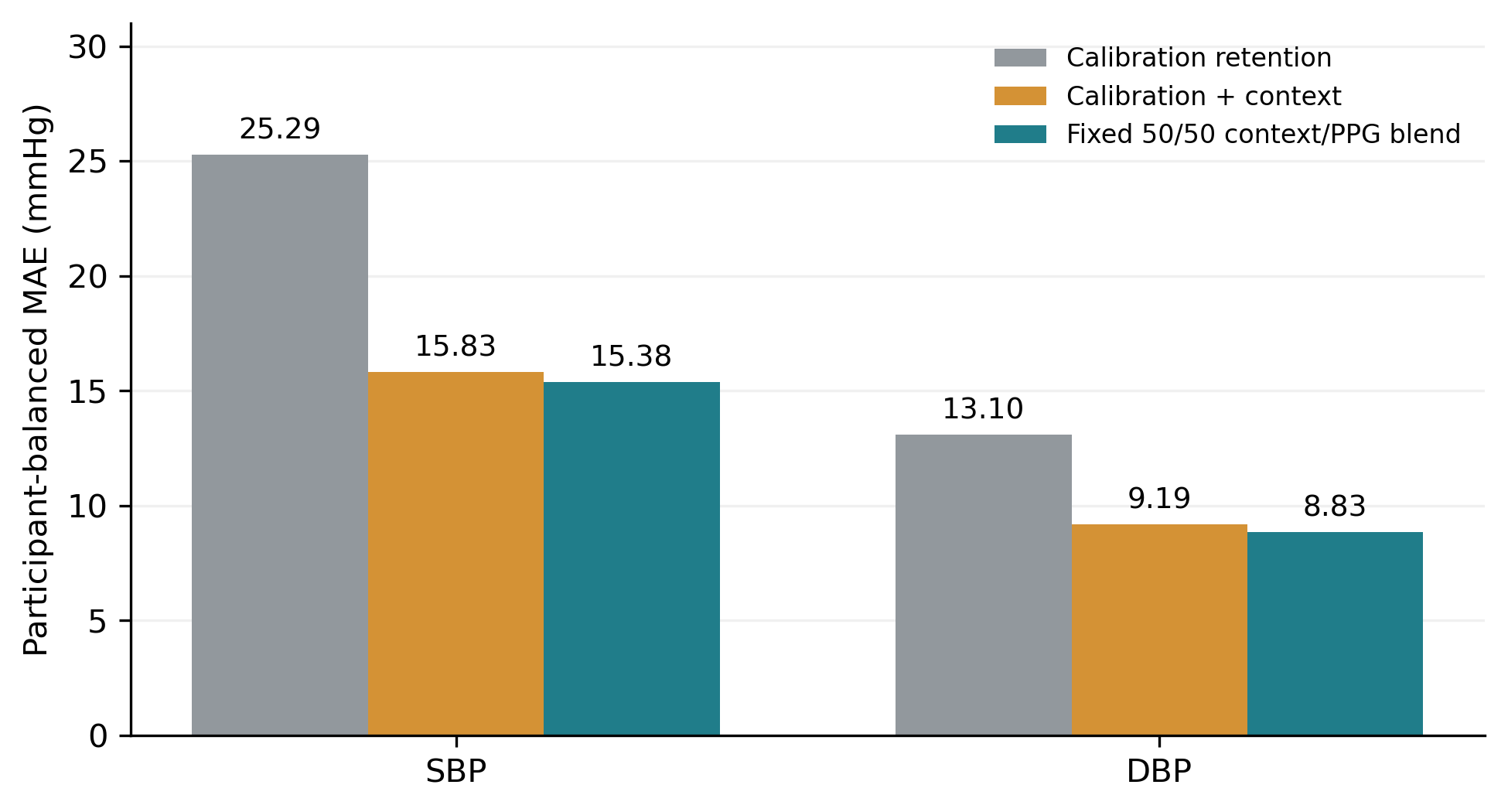}
\caption{VitalDB locked-holdout participant-balanced MAE. Calibration-plus-context explains most of the gain over retention; the frozen fixed context/PPG blend adds small CI-supported incremental improvements for both endpoints.}
\label{fig:vitaldbpilot}
\end{figure}

Figure~\ref{fig:vitaldbpilot} and Table~\ref{tab:vitaldbpilot} distinguish retention from the stricter calibration-plus-context control. Most improvement came from context. The fixed 50/50 blend reduced SBP/DBP MAE from $15.827$/$9.189$ to $15.375$/$8.833$~mmHg; blend-minus-context differences were $-0.451$ (95\% CI $-0.781$--$-0.121$) and $-0.357$ ($-0.512$--$-0.194$). It improved on context for 67.0\%/70.5\% of subjects. For reference changes of at least 10~mmHg, gains were $0.707$/$0.501$~mmHg (intervals in Supplementary Section E). Absolute errors remain high; this retrospective intraoperative result establishes neither clinical adequacy nor prospective wearable performance.
\begin{table}[t]
\centering
\caption{VitalDB frozen holdout (227 subjects). Fixed blend is the arithmetic mean of context and PPG-augmented predictions; negative blend-minus-context differences favor the blend. All entries are participant-balanced MAEs in mmHg.}
\label{tab:vitaldbpilot}
\scriptsize
\resizebox{\columnwidth}{!}{%
\begin{tabular}{lrrrr}
\toprule
Target & Retention & Context & Fixed blend & Blend$-$context [95\% CI] \\
\midrule
SBP & 25.291 & 15.827 & 15.375 & $-0.451$ [$-0.781$, $-0.121$] \\
DBP & 13.098 & 9.189 & 8.833 & $-0.357$ [$-0.512$, $-0.194$] \\
\bottomrule
\end{tabular}}
\end{table}
\subsection{State-of-the-Art Comparison}
\label{sec:sota}
Table~\ref{tab:vitalbenchmark} supplies a main-text comparison of relevant calibrated intraoperative studies; Supplementary Table S9 retains the broader comparison. Published values are not head-to-head reruns, and different populations, selection rules, calibration schemes and averaging procedures prevent a direct numerical ranking.

\begin{table}[t]
\centering
\caption{Calibrated intraoperative PPG studies. Errors are MAE in mmHg. Source-specific calibration and evaluation conditions are essential to interpretation; SNU denotes Seoul National University Hospital.}
\label{tab:vitalbenchmark}
\small
\begin{tabular}{p{3.3cm}p{7.0cm}rr}
\toprule
Study & Data and calibration/evaluation protocol & SBP & DBP \\
\midrule
Joung et al.\ \cite{ppg2bp2023} & SNU surgical cohort, partly represented in VitalDB; paired calibration/inference network; participant-independent development/test split. Whole 797-subject test set. & 7.991 & 4.789 \\
Joung et al.\ ABP-20m subset & 629 subjects assessed after 20 minutes following arterial-line insertion; subset of the preceding cohort. & 5.525 & 3.282 \\
Dias et al.\ \cite{dias2024preprint} & VitalDB; 1,068 eligible patients; normalized PPG/calibration-window pairs 3 min--2 h apart, $|\Delta\mathrm{SBP}|\leq60$~mmHg; unfiltered PPG result from author-posted Table 3. & 9.20 & 5.68 \\
Kim and Baek \cite{kim2026transfer} & VitalDB target; cross-dataset pretraining followed by target-subject fine-tuning. Personalization uses 50--1,800 five-second segments; main comparison uses 1,800, with 720 validation and 720 test segments per subject. & 3.53 & 1.96 \\
This study & VitalDB; 55 development / 227 disjoint held-out subjects; three early arterial-reference windows; fixed blend evaluated on 5,472 future windows, with equal subject weighting. & 15.375 & 8.833 \\
\bottomrule
\end{tabular}
\end{table}

The present VitalDB errors are larger than these published calibrated estimates. The Joung cohort is not identical to the public VitalDB release, and its ABP-20m result is a selected subset; its reported mean error $\pm$ SD must not be confused with the MAEs shown here. Dias et al.'s numerical row is explicitly taken from the author-posted preprint, rather than assumed to reproduce the final publication \cite{limitations2024}; its pairing and BP-change restrictions differ from an initial-window anchor. Kim and Baek's target-subject parameter fitting entails substantially more reference data than a three-window anchor. These differences qualify interpretation but do not establish that the present model performs well. No comparator has been rerun on our 227-subject holdout, so relative superiority or inferiority under an identical protocol remains untested.

\subsubsection{VitalDB error structure and practical interpretation}
\label{sec:vitaldiagnostics}
The unchanged frozen-blend predictions have participant-balanced bias of $-2.359$~mmHg SBP and $-2.057$~mmHg DBP, with individual-error SDs of $19.760$ and $11.312$~mmHg. Mean absolute reference departures from the initial anchor are $25.291$ and $13.098$~mmHg. Exploratory stratification of these saved predictions is reported in Supplementary Table S12. For SBP reference values below 110, from 110 to below 140, and at least 140~mmHg, participant-balanced MAEs within each stratum are $15.559$, $9.436$ and $34.104$~mmHg, respectively. Corresponding DBP MAEs below 60, from 60 to below 80, and at least 80~mmHg are $7.694$, $6.174$ and $22.845$~mmHg. The same subject can contribute to multiple strata, so these stratum means should not be averaged to reconstruct overall MAE. Errors are also high for departures of at least 20~mmHg from the initial anchor ($20.407$~mmHg SBP and $17.220$~mmHg DBP).

These descriptive patterns identify limitations at high pressures and large changes from calibration. A small development cohort, the chosen feature representation and estimator capacity, signal/reference alignment, and perioperative physiological variation are possible contributors; this analysis does not identify their causal effects. No high-error observations were removed and no holdout-based retuning was performed for this revision. The practical evidence is restricted to a small incremental PPG contribution after a strong calibration-plus-context control. It does not justify BP-based clinical decisions, clinical interchangeability, or a claim that the absolute errors are satisfactory.

\subsection{Exploratory Sensor-History Extension}
\label{sec:followupresults}
Table~\ref{tab:followupmodels} retains all three follow-ups to the static-rich reference. The enriched median/calibration ensemble reduced MAE by only $0.017$/$0.008$~mmHg SBP/DBP, with both 97.5\% intervals including zero. Direct absolute-BP prediction increased MAE. Adding recording-level sensor history gave the lowest observed MAEs in this series: $7.299$/$5.445$~mmHg, gains of $0.064$ (97.5\% interval $0.021$--$0.108$) and $0.025$ ($0.007$--$0.045$) relative to the static-rich reference.

The matched fixed-weight static-versus-history comparison also favored history, by $0.058$~mmHg SBP (descriptive 95\% interval $0.022$--$0.096$) and $0.018$~mmHg DBP ($0.004$--$0.033$). This contrast isolates history under fixed weights; comparison with the inner-selected static-rich reference also changes the blending policy. The result does not isolate a nonlinear/attractor feature effect and is not a matched improvement over the original primary configuration.

\begin{table}[t]
\centering
\caption{Later exploratory AuroraBP models: participant-balanced MAE (mmHg), with the same cohort, initial three-reading burden and participant folds within this series. The static-rich reference is separate from the original primary model in Table~\ref{tab:v26primary}. Its features and blending differ from that primary model. Specifications were frozen per experiment after earlier dataset-guided development.}
\label{tab:followupmodels}
\small
\begin{tabular}{lrr}
\toprule
Model & SBP MAE & DBP MAE \\
\midrule
Static-rich reference & 7.363 & 5.470 \\
Median/calibration ensemble & 7.346 & 5.462 \\
Direct-BP fixed ensemble & 7.662 & 5.567 \\
Causal sensor-history ensemble & 7.299 & 5.445 \\
\bottomrule
\end{tabular}
\end{table}

The history model's signed bias was $-0.122$/$0.050$~mmHg and marginal error SD was $10.379$/$8.032$~mmHg. SBP absolute errors above 20~mmHg decreased from $6.45\%$ to $6.25\%$ under participant balancing, whereas DBP errors above that threshold increased from $2.50\%$ to $2.59\%$. Thus the small MAE reduction does not imply uniformly better individual accuracy. Supplementary Tables S14--S17 report the paired contrasts, dispersion, subgroups and remaining error structure.

\section{Discussion}
\label{sec:discussion}
The primary comparison supports adding wearable-state context to its calibrated core, which already contains optical/time candidates. The matched factorial supplies complementary no-PPG controls and shows small conditional optical contributions. Removing clock terms reduces one protocol shortcut, but context still uses supplied posture/activity. No independently estimated wearable-state model has been evaluated, so annotation gains cannot establish deployment generalization.

The studies answer distinct questions. Quality weighting improved pulse-rate extraction relative to equal weighting. VitalDB showed a small incremental contribution from a separately fitted PPG implementation, combined with context through a fixed 50/50 blend. It tested neither quality-dependent blending nor transfer of AuroraBP parameters. The later sensor-history series improved MAE modestly within its own matched folds, while direct BP targets worsened error and DBP tails did not improve uniformly. These results identify conditional predictive information, not an attractor-specific mechanism or a universally generalizing architecture.

Near-zero aggregate bias coexists with broad individual errors. The historical measurement-level SBP standards comparison yielded BHS Grade C and error SD $9.718$~mmHg, exceeding the cited $8$~mmHg numerical cutoff; VitalDB MAE remained $15.375$/$8.833$~mmHg. Neither relative gains nor exclusion of difficult predictions can establish clinical adequacy. Pigmentation interventions likewise support no fairness conclusion. The contribution is a transparent calibrated evaluation with retained negative findings, rather than demonstrated cuff replacement or clinical outcome benefit.
\section{Limitations}
\label{sec:limitations}
\textbf{Validation.} AuroraBP is a controlled single-site protocol. Although initial visits are shorter than one hour, return targets occur 21.051--32.853~h after initial calibration. Participant-separated folds prevent direct participant overlap but cannot remove optimism from earlier dataset-guided choices. Appel--Theart evaluates pulse rate; VitalDB is a separate retrospective intraoperative model. Neither establishes prospective wearable BP performance or transfer of one model.

\textbf{Calibration and context.} Primary tracking requires three initial cuff readings. The one-reading protocol-enriched arm has higher MAE ($8.257$/$6.280$~mmHg); a matching wearable-state dose analysis is unavailable. Protocol annotations and enriched metadata may encode acquisition structure. The history ensemble retains time and protocol-dependent sampling; a matched time-free version and new-protocol validation are unavailable. History checks establish recording-level order, not within-window latency or real-time operation.

\textbf{Pigmentation.} Only 25 primary and six historical V--VI participants are represented. Auxiliary pigmentation ranks are synthetic. Null/adverse results for one correction establish neither equivalence nor impossibility of better interventions. Pulse fusion does not validate BP equity, and VitalDB has no pigmentation labels.

\textbf{Inference and reproduction.} Holm adjustment covers the two retrospective primary endpoints; exploratory intervals do not account for adaptive experiments, overlapping training sets or method selection. Repeated readings are handled at participant level, but agreement assumptions can fail. The DBP router's $2.9\%$ recall precludes a safety claim. BHS/AAMI/ISO summaries are numerical context, not certification. Re-extracting all 11,427 standard AuroraBP rows reproduced the preserved extractor's cache (relative tolerance $10^{-6}$, absolute $10^{-10}$); this checks consistency, not independent algorithm validity, and excludes rich/beat extraction. The newer-runtime bridge verifies one original fold, not every fit. Complete third-party raw-to-results replication remains outstanding. Supplementary Sections A, D and E detail these boundaries.
\section{Conclusion}
\label{sec:conclusion}
Matched calibrated controls and participant-separated evaluation showed small PPG/context contributions, with the DBP PPG increment sensitive to time removal. Nonlinear augmentation had no established additional benefit. Sensor history gave small exploratory gains relative to its own reference, without uniformly improving DBP tails. Independent analyses supported pulse-rate fusion and a small PPG contribution in a separate VitalDB model. The traceable workflow and retained negative findings do not establish clinical adequacy, model transfer, equitable performance or cuff replacement.
\section*{Code and Reproduction Availability}
The original companion archive is public:
\begin{center}\small\url{https://github.com/farouze/pc-avct-reproducibility}\end{center}
Its versioned release contains the original primary analysis, feature-family ablation, helpers, dependencies, hashes, manifests and aggregate outputs. Reviewer extensions, historical reconstructions and the static-rich/three-follow-up chain are supplied in the accompanying revision package; they have not been uploaded to that repository. Restricted feature banks, participant predictions and contrasts remain local under dataset access conditions.

Feature-cache-to-results reproduction requires the specified hashed inputs and recorded environment. Comparator predictions, source equivalence, history ordering and aggregate inference were checked. The standard AVCT bank also has a separate waveform re-extraction check, but rich/beat extraction was not independently regenerated and complete third-party replication remains outstanding. Full follow-up fits used equivalent Python sources/local workers; notebook definitions were checked, and recorded summaries are distinguished from executed notebook outputs. Supplementary Section D.4 gives execution order and scope. No software DOI is claimed; dataset DOIs identify data.

\section*{Data Availability}
AuroraBP is available from its provider upon application through the Zenodo record cited in \cite{aurorabpDataset} (DOI: 10.5281/zenodo.19099166) and acceptance of the Data Use Agreement. The authors are not permitted to redistribute AuroraBP participant-level data. BIDMC is available through PhysioNet. MIMIC-II is a historical development source; its cited clinical-database version is no longer distributed by PhysioNet. UCI Cuff-Less BP is available through the UCI Machine Learning Repository, the multi-wavelength data through Mendeley Data \cite{appel2025ppg}, and VitalDB through PhysioNet \cite{vitaldbDataset}. Access conditions and citations appear in Section~\ref{sec:datasets}. No directly identifying participant data are included in the manuscript or figures.
\section*{Ethics Statement}
This study is a secondary computational analysis of previously collected, deidentified human data. The authors recruited no participants, performed no intervention, had no participant contact, received no direct identifiers, and made no attempt to reidentify participants. Morgan State University's published IRB process states that submission is not required for research using deidentified data or publicly available datasets \cite{morganIRB2023}. Accordingly, this analysis did not require review by the Morgan State University Institutional Review Board, and no approval number was issued. The analysis followed the ethical principles of the Declaration of Helsinki and applicable local requirements. Ethics approvals and written informed-consent procedures for the original data collections are reported by the respective dataset providers and source publications \cite{aurorabp2022,aurorabpDataset,pimentel2017,appel2025ppg,lee2022vitaldb,vitaldbDataset}. No additional consent or consent for publication was required because the authors used only deidentified secondary data and report no identifiable participant information. This study was not a clinical trial.
\section*{Author Contributions}
Farouk Ganiyu Adewumi: conceptualization, methodology, software, validation, formal analysis, data curation, visualization, and writing of the original draft. Timothy Oladunni: conceptualization, methodology, supervision, and review and editing of the manuscript.
\section*{Competing Interests}
The authors declare no competing financial or non-financial interests.
\section*{Funding}
This work was supported in part by Morgan State University's Office of Research and Graduate Programs.
\section*{Acknowledgment}
We thank the participants, investigators, and data stewards who made the secondary datasets used in this study available.
\begingroup
\small
\interlinepenalty=10000
\bibliographystyle{pmea_harvard}
\bibliography{references}
\endgroup
\end{document}